\documentclass[10pt, a4paper, conference, compsocconf]{IEEEtran}
\usepackage{CJKutf8}
\usepackage{xcolor}
\usepackage{graphicx}
\usepackage{url}
\usepackage{amsmath}
\DeclareMathOperator*{\argmax}{arg\,max}

\ifCLASSINFOpdf
\else
\fi
\begin{document}
%
\title{Bare Demo of IEEEtran.cls for IEEECS Conferences}


\author{\IEEEauthorblockN{Chia-Yu Li}
\IEEEauthorblockA{Institute for Natural Language Processing (IMS)\\
University of Stuttgart\\
Stuttgart, Germany\\
licu@ims.uni-stuttgart.de}
\and
\IEEEauthorblockN{Ngoc Thang Vu}
\IEEEauthorblockA{Institute for Natural Language Processing (IMS)\\
University of Stuttgart\\
Stuttgart, Germany\\
nthangvu@ims.uni-stuttgart.de}
}


%


\title{Integrating Knowledge in End-to-End Automatic Speech Recognition for Mandarin-English Code-Switching}
\maketitle

\begin{abstract}
Code-Switching (CS) is a common linguistic phenomenon in multilingual communities that consists of switching between languages while speaking. This paper presents our investigations on end-to-end speech recognition for Mandarin-English CS speech. We analyze different CS specific issues such as the properties mismatches between languages in a CS language pair, the unpredictable nature of switching points, and the data scarcity problem. We exploit and improve the state-of-the-art end-to-end system by merging nonlinguistic symbols, by integrating language identification using hierarchical softmax, by modeling sub-word units, by artificially lowering the speaking rate, and by augmenting data using speed perturbed technique and several monolingual datasets to improve the final performance not only on CS speech but also on monolingual benchmarks in order to make the system more applicable on real life settings. Finally, we explore the effect of different language model integration methods on the performance of the proposed model. Our experimental results reveal that all the proposed techniques improve the recognition performance. The best combined system improves the baseline system by up to 35\% relatively in terms of mixed error rate and delivers acceptable performance on monolingual benchmarks.
\end{abstract}


%
\IEEEpeerreviewmaketitle

\section{Introduction}
Code-switching (CS) speech is a common phenomenon in multilingual countries and defined as speech which contains more than one language \cite{b1}. From a grammatical point of view, Poplack \cite{b2} proposed three types of CS: extra-sentential, inter-sentential and intra-sentential. Extra-sentential switching is inserting tag elements from  one language into an otherwise monolingual language. Inter-sentential switching is characterized by a switch from one language to another outside the sentence or the clause level, whereas intra-sentential switching is switching from one language variety to another at the clause, phrase, or word level within a single utterance. This paper aims at improving end-to-end (E2E) automatic speech recognition (ASR) system on the SEAME corpus (South East Asia Mandarin-English) \cite{b3} which is intra-sentential dominant \cite{b4}. 

A first ASR system for Mandarin-English CS conversational speech was proposed in \cite{b5} investigating different merged acoustic units for acoustic modeling, artificial CS data for language modeling, and the use of language identification in the decoding process. Recent studies show that deep learning has boosted the performance of ASR \cite{b6,b7} and the state-of-the-art ASR architecture - hybrid TDNN-HMM - has shown incredible performance on many LVCSR tasks \cite{b9}. Despite hybrid ASR having state-of-the-art performance, building this system remains a complicated and expertise-intensive task. First, it requires various resources such as pronunciation dictionaries and phonetic questions for acoustic modeling. Second, it relies on GMMs for frame-level alignments. In the context of CS, creation of a pronunciation dictionary for two languages might require expertise knowledge, e.g., generating pronunciation variants, and this process is error prone. 
Recently, some studies proposed a single neural network architecture to perform speech recognition in an end-to-end manner to resolve the issues in hybrid ASR. There are two types of E2E frameworks: CTC based \cite{b10,b14} and attention based \cite{b15,b18}. Although the attention model has been shown to improve the performance over CTC based, it is difficult to learn in the initial training stage with long input sequences and has poor performance in noisy conditions. Joint CTC-attention based E2E framework was proposed to improve noise robustness, achieving fast convergence and mitigating the alignment issue \cite{b20}. The experimental results on many benchmarks (WSJ, CHiME-4, etc.) demonstrate its advantages over both the CTC and attention based frameworks and comparable results to hybrid ASR systems.

\begin{CJK*}{UTF8}{gbsn}
\indent Mandarin and English have many significant differences \cite{b21,b22}. First, Mandarin uses a logographic system, in which symbols represent the words' meaning and not their pronunciation. Second, Mandarin is a tone language that uses the pitch to distinguish word meaning, whereas English uses the pitch to express emotion or emphasize words. Third, the syntactic structure. e.g. in English, things are usually modified by the words that come after them, while in Mandarin, things are usually modified by the words that precede them. Furthermore, it is difficult to predict the CS points which is entirely up to the individual speakers \cite{b2}. Take the below sentences S1 and S2 as examples, the English word 'GO' has similar pronunciation as the Mandarin character '够', but both have totally different meanings. Both sentences can possibly occurr in CS environment. If the ASR system is mainly trained with monolingual Mandarin data, then it is more likely to predict the next character to be '够' given the history '我'. Some might argue that the ASR systems can learn the conditional probability $P(GO\vert$ 我$)$ from the CS data. However, collecting CS data is time-consuming and financially expensive. Besides, the CS points highly depend on the speakers \cite{b2}, and it is not easy to cover all possible CS points.
\begin{itemize}
    \item S1: 我 \hspace{1mm} GO \hspace{1mm}了 (Translation: I go.)
    \item S2: 我 \hspace{1mm} 够 \hspace{4mm}了 (Translation: It is enough for me.)
\end{itemize}
\end{CJK*}
In this paper, we analyze several issues of Mandarin-English CS speech which might cause recognition errors and inject knowledge derived from the analysis into the development process of the E2E ASR system. We merge discourse particles and nonlinguistic signals, integrate language identification into the prediction process, utilize English subword modeling, artifically lower speaking rate, and use data augmentation to solve these issues. Furthermore, we investigate the effect of these techniques and their combinations on the ASR performance in terms of Mixed Error Rate (MER). Finally, we explore different language model integration methods in order to interpolate the knowledge of the language model into our best combined systems.

\section{SEAME dataset}
SEAME is a 99 hours of spontaneous Mandarin-English CS speech corpus recorded from Singaporean and Malaysian speakers. All recordings are performed by close-talk microphone in quiet room. The speakers are aged between 19 and 33, almost balanced in gender (49.7\% of female and 50.3 \% of male). The total number of distinct speakers is 157 (36.8\% are Malaysian while the rest are Singaporean) \cite{b24}. 16.96\% of utterances are English (ENG), 15.54\% are Mandarin (MAN) and the rest (67\%) are CS utterances. In each transcript, they use the following categories for labeling: target language (English word and Mandarin character), discourse particle or hesitation ('lah', 'hmm', etc.) and nonlinguistic signal (people laughing, coughing, etc.), other languages (Japanese or Korean words).

\section{Identification of subproblems}
\subsection{Discourse particle and nonlinguistic signal}
\label{sec:nlsyms}
There are 430 unique discourse particles, hesitations and nonlinguistic signals (people laughing, coughing). These signals might be informative for sentiment analysis or emotion detection not for speech recognition. 

\subsection{Code-Switching points prediction}
\label{sec:csp}
There are two language switching directions: one is from English to Mandarin and another is from Mandarin to English. SEAME has 12.24\% switching points (6.04\% are from English to Mandarin and 6.2\% are from Mandarin to English). Previous studies state that the code-switching points are indeterminate because the code-switching decision is entirely up to the individual speakers \cite{b25} and there are some code-switching patterns across speakers \cite{b2}. 
Moreover, in over 80\% of cases, speakers directly switch language without any short pause and discourse particle between two adjacent different languages \cite{b3}. It is a challenge for conventional ASR system to predict the switching points due to insufficient acoustic information. 

\subsection{Out-of-Vocabulary (OOV)}
\label{sec:oov}
OOV is a common problem in the context of speech recognition and would be accumulated due to the recognition of two languages. For example, there are around 370,000 Mandarin Chinese words and 172,000 English words. If we just combine two dictionaries to a lexicon for the ASR systems, the tedious lexicon would make ASR hard to be trained due to huge memory and time consumption. Not to mention it does not contain the new words being created in daily life or social media. 

\subsection{High speaking rate}
\label{sec:highspeakingrate}
A rate of clear speech ranges between 140-160 words per minute (wpm) and a rate higher than 160 wpm can make it difficult for the listener to absorb the material. Reference \cite{b3} reports that Singaporean speakers have an average speaking rate of 181 wpm and Malaysian speakers 151 wpm. Note that there are 72 hours of speech from Singapore, and 27 hours from Malaysia. Therefore, around 70\% of the utterances have high speaking rates.

\subsection{Data scarcity}
\label{sec:datascarcity}
Although there are many multilingual countries, only few countries do CS between Mandarin and English. Besides, CS speech normally occurs in casual conversation and it is not possible to record it for free due to the privacy concern. Especially in the context of E2E ASR, data scarcity might be a large problem to build a good system.

\section{Proposed methods}
\subsection{End-to-End speech recognition}
It has been shown that the joint CTC-attention model within the multi-task learning framework \cite{b20} is able to outperform CTC-based or attention-based E2E ASR systems due to its robustness, fast convergence, and mitigation of the alignment issues. Furthermore, it allows building ASR systems without the use of a pronunciation dictionary, which is convenient for CS ASR because combining two languages' pronunciation dictionaries requires expertise knowledge.
The overall architecture contains the shared encoder which is trained with both CTC and attention model objectives simultaneously and transforms the input sequence \textbf{\textit x} into high level features \textbf{\textit h}, and the location-based attention decoder generates the character sequence \textbf{\textit y} \cite{b28}. The multi-task learning (MTL) objective, is represented in Eq.\ref{eq1}, follows by using both CTC and attention model.

\begin{equation}
\footnotesize
  \mathcal{L}_{MTL} = \lambda \mathcal{L}_{CTC} + (1-\lambda) \mathcal{L}_{Attention}
  \label{eq1}
\end{equation}
where \(\mathcal{L}_{CTC}\) is the loss function of the CTC model, \(\mathcal{L}_{Attention}\) is the loss function of the attention model, and a tunable parameter $0 \leq \lambda \leq 1$.

\subsection{Merging discourse particles and nonlinguistic signals}
In \ref{sec:nlsyms}, we mention the huge amount of labels for discourse particles, hesitations, and nonlinguistic signals. The system should put effort on learning language instead of nonlinguistic symbols.  
Therefore, we group all the discourse particles and hesitation pauses into the same class, e.g., "lah" and "hmm" are labeled as "$<$dispar$>$", and all the nonlinguistic signals are labeled as "$<$nlsyms$>$". The goal is to let neural network focus on learning language (English and Mandarin) characters because they will be the main factors to the loss function. 

\subsection{Language identification using hierarchical softmax}
To predict the CS points mentioned in \ref{sec:csp}, we exploit language identification to predict if the current word (character) is English or Mandarin given the history in terms of the high level features $h$ (or the output of Encoder). The language identification is integrated into E2E attention model with the output layer factorized by class layer (hierarchical softmax), proposed in \cite{b29}. The probability of character at the $i$-th time step $y(i)$ given \textit{history} is defined as
\begin{equation}
\footnotesize
  P(y(i)|history) = P(s(i)|history)P(y(i)|s(i))
  \label{eq4}
\end{equation}
where $s(i)$ denotes the type of language (English or Mandarin) at the $i$-th time step. 

Furthermore, the overall E2E system is trained using multi-task learning objective represented in Eq.\ref{eq5} using CTC, attention, and language identification models.
\begin{equation}
\footnotesize
  \mathcal{L}_{MTL} = \lambda \mathcal{L}_{CTC} + (1-\lambda) (\mathcal{L}_{Attention} + \mathcal{L}_{ld})
  \label{eq5}
\end{equation}

\subsection{English subword modeling}
\ref{sec:oov} identifies the OOV problems for both languages. For Mandarin, we could use characters as unit to mitigate the OOV problem because our E2E system only needs to recognize 50,000 characters instead of 370,000 words. For English, the subword model can solve the OOV problem and offer a capability in modeling longer context than using characters \cite{b30,b42}.

\subsection{Lower speaking rate}
In \ref{sec:highspeakingrate}, we mentioned the problem of high speaking rate causing the difficulty for recognition. In a daily conversation, when people do not understand what others say, normally they ask others to say it again, and they will repeat it with a lower speaking rate. Motivated by this observation, we propose to artificially lower the speed of the entire dataset.

\subsection{Data augmentation}
We use `3 way speed-perturbed' method proposed in \cite{b31} to generate more CS data. In particular, two additional copies of the original training data are generated by modifying the speed to 0.9 and 1.1 of the original rate and added to the original data.
Furthermore, we add several monolingual datasets for the training process allowing our system to learn more pronunciation variants.

\subsection{Language model integration methods}
There are several ways to integrate E2E ASR systems with an external language model. In a conventional decoding paradigm with an external language model, shallow fusion (SF) computes the score by linearly interpolating the score from a Sequence-to-Sequence (S2S) model and an external language model to maximize the following criterion:
\begin{equation}
\footnotesize
    y^{*} = \argmax_{y \in \Omega^{*}}\{\ln P_{S2S}(y|x)+\beta P_{LM}(y)\}
\end{equation}
where $x$ is acoustic features and $y$ is the sequence made of English words and Mandarin characters. Where $\beta$ is a tuneable parameter to define the importance of the external LM. 
Unlike SF uses the language model in the decoding state, cold fusion (flat-start fusion) (CF) uses the pre-trained language during the training of the S2S model to provide effective linguistic context \cite{b41}. The fine-grained element-wise gating function is equipped to flexibly rely on the language model depending on the uncertainty of predictions:
\begin{equation}
\footnotesize
    s_{t}^{LM}=DNN(d_{t}^{LM})
\end{equation}
where $d_{t}^{LM}$ is the hidden states of RNNLM, $s_{t}^{LM}$ is a feature from the external LM. The S2S models' hidden states $s_{t}^{ED}$ is defined as:
\begin{equation}
\footnotesize
    s_{t}^{ED}=\sigma(W_{ED}[d_{t};c_{t}]+b^{ED})
\end{equation}
CF uses a fine gating mechanism, and the gating function $g_{t}$ takes features from the S2S model and the external LM.
\begin{equation}
\footnotesize
    g_{t}=\sigma(W^g[s_{t}^{ED};s_{t}^{LM}]+b_g)
\end{equation}
\begin{equation}
\footnotesize
    s_{t}^{CF}=[s_{t}^{ED};g_{t} \circ s_{t}^{LM}]
\end{equation}
where $\circ$ is element-wise multiplication.

\begin{equation}
\footnotesize
    r_{t}^{CF}=DNN(s_{t}^{CF})   
\end{equation}
\begin{equation}
\footnotesize
    P(y_{t}|h,y<t)=softmax(W_{CF}r_{t}^{CF}+b^{CF})
\end{equation}


\section{Experimental setup}
The corpus is split into train, development, and evaluation sets. The statistics of the three sets are shown in Table \ref{tab:data}. CS, MAN, and ENG represent Code-switching, Mandarin, and English utterances, respectively. Based on the ratio of CS, MAN, and ENG utterances in the three sets, the evaluation set is Mandarin dominant and the development set is relatively bilingual balanced.\\ 
\tabcolsep=0.11cm
\begin{table}[!thb]
\scriptsize
  \caption{\scriptsize The Statistics of the train, dev and eval sets}
  \label{tab:data}
  \centering
  \begin{tabular}{|l|c|c|c|c|c|r|}
    \hline
    \multicolumn{4}{|c}{}&\multicolumn{3}{|c|}{\textbf{Ratio(\%)}} \\
    \hline
    \textbf{Sets} & \textbf{\# spk} & \textbf{\# utt} & \textbf{\#hrs} & \textbf{CS} & \textbf{MAN} & \textbf{ENG}\\
    \hline
    train   & 141 & 93782 & 96 & 67.19 & 15.45 & 17.36 \\
    dev     &   8 & 6549 & 1.8  & 66.63 & 17.03 & 16.34 \\
    eval    &   8 & 5037 & 1.4  & 74.08 & 15.53 & 10.38 \\
    \hline
  \end{tabular}
\end{table}

\subsection{Baseline system}
The baseline system is trained with the most current recipe from Espnet \cite{b28}. The encoder network is represented by bidirectional long  short-term memory (BLSTM) with subsampling and has 5 layers with 1024 units. The decoder is represented by 1 layer of BLSTM with 1024 units. The hybrid CTC/attention parameter ($\lambda$) (Eq.\ref{eq1}) is $0.5$. The beam size is 20 and the CTC weight is 0.5 for decoding. The dictionary is character based.

\subsection{Additional monolingual data}
For Mandarin Chinese dataset, we utilize Aishell-1 which contains 170 hours of speech contributed by 400 people from different accent areas in China \cite{b34}, THCHS30 containing 30 hours of Mandarin Chinese speech database \cite{b35}, and Free ST Chinese Mandarin Corpus (ST-CMDS) having 110 hours (855 speakers) of speech recorded in a silent indoor environment using a cellphone \cite{b36}.
The English datasets in the experiment are 1000 hours of Librispeech \cite{b37} and 425 hours of Common Voice \cite{b38}, and 5 hours of Ted talks extracted from the TEDxSingapore website\cite{b39}.

\section{Results \& analysis}
\label{sec:results_analysis}
This section presents a performance comparison in terms of MER (\%) between the baseline (joint CTC/attention E2E) and all proposed solutions that are denoted as E2ELD (baseline with language identification), E2ESW (baseline with subword modeling), SL (slowing down the speed of utterance), 3W (3-way speed perturbation), and F (adding monolingual data).
Label 1 denotes the dataset using the original labels and label 2 denotes the dataset using our proposed labels (merging discourse particles and nonlinguistic signals). Moreover, we examine the systems on the test sets with and without nonlinguistic symbols (discourse particles and nonlinguistic signals). The test set without nonlinguistic symbols will show how well the system recognizes actual language. 

\subsection{Merging discourse particles and nonlinguistic signals, and language identification}
Table \ref{tab:B} shows that the baseline system trained with label 2 does not outperform the one trained with label 1. However when integrating language identification information in the output layer using hierarchical softmax, the system trained with label 2 data improves the performance especially on the test sets without nonlinguistic symbols (No nlsyms). The hypothesis of baseline E2E and E2ELD for one utterance in eval set are shown in Table \ref{tab:expB}. The output of baseline has mistakenly recognize English word 'initiative' as the sequence of English characters and Mandarin characters in blue ink while E2ELD model has better identification between languages.

\begin{table}[htbp]
\scriptsize
  \caption{\scriptsize The MER(\%) on SEAME test set of baseline and all proposed models}
  \label{tab:B}
  \begin{center}
  \begin{tabular}{|l|c|cc|cc|}
    \hline
    &\textbf{Label}&&&\multicolumn{2}{c|}{\textbf{No nlsyms}}\\
    \hline
    \textbf{Systems}&\textbf{type}&\textbf{dev}&\textbf{eval}&\textbf{dev}& \textbf{eval} \\
    \hline
    E2E & 1 & 39.8 & 31.7 & 39.8 & 31.0 \\
    E2ELD        & 1 & 39.1  & 31.4 & 38.7 & 30.6\\
    E2ESW(SEAME500) & 1 & 34.6  & 27.9 & 34.1 & 27.1\\
    E2E+SL(0.7) & 1 & 36.5  & 29.2 & 36.1 & 28.4\\
    E2E+3W        & 1 & 34.8  & 27.7 & 34.4 & 26.8\\
    \hline
    E2E & 2 & 40.3 & 31.6 & 41.1 & 31.4 \\
    E2ELD        & 2 & 38.1  & 29.3 & 38.7 & 28.9\\
    E2ESW(SEAME500) & 2 &33.6&26.4&33.9&25.7  \\
    E2E+SL(0.8) & 2 &37.3&29.0&38.0&28.6  \\
    E2E+3W        & 2 & 34.4  & 26.7 & 34.3 & 26.2\\
    \hline
  \end{tabular}
\end{center}
\end{table}

\tabcolsep=0.11cm
\begin{CJK*}{UTF8}{gbsn}
\begin{table}[htbp]
\scriptsize
  \caption{\scriptsize The hypothesis of baseline and E2ELD and E2E-SL models for one utterance in eval set}
  \label{tab:expB}
  \begin{center}
  \begin{tabular}{|l|l|}
  \hline
  {\bf Systems} &{\bf Hypothesis}\\
  \hline
    Ground-truth & then \hspace{1mm}你 \hspace{1mm}不 \hspace{1mm}可 \hspace{1mm}以 \hspace{1mm}take initiative\hspace{1mm} 去\hspace{1mm}讲 \hspace{1mm}么\\
    E2E  & then \hspace{1mm}你 \hspace{1mm}不 \hspace{1mm}可 \hspace{1mm}以 \hspace{1mm}\textcolor{red}{that} \textcolor{blue}{in} \hspace{1mm}\textcolor{blue}{你}\hspace{1mm}\textcolor{blue}{学}\hspace{1mm}\textcolor{blue}{tive}\hspace{1mm}\textcolor{red}{就} \hspace{1mm}\textcolor{red}{讲} \hspace{1mm}\textcolor{red}{嘛} \\
    E2ELD & then \hspace{1mm}你 \hspace{1mm}不 \hspace{1mm}可\hspace{1mm}以\hspace{1mm}\textcolor{red}{tat}  initiative\hspace{1mm}\textcolor{red}{就}\hspace{1mm}讲 \\
    \hline
    Ground-truth & why you want to be the head of your of your group of friends\\
    E2E  & why \hspace{4mm} want to be the head \hspace{8mm} of your group of friends\\
    E2E-SL (0.7) & why you want to be the head of your of your group of friends\\
  \hline
  \end{tabular}
  \end{center}
\end{table}
\end{CJK*}

\subsection{English subword modeling}
We use two different texts (SEAME and Librispeech) to train the English subword model and add different amounts (100$\sim$5000) of subwords to the dictionary. The result shows that E2ESW with 500 subwords trained from SEAME text has better performance than the one with 500 subwords trained from Librispeech. The reason is that the frequent subwords in Librispeech and SEAME are not similar. To be more specific, SEAME has many conversation style English words and proper names related to South Asia while Lbrispeech mainly contains literary words. Therefore, the words in Librispeech are not likely to be used in a casual conversation in South Asia.


\subsection{Lower speaking rate}
We examine different factors (0.6$\sim$0.9) to lower the audio speed.  Table \ref{tab:B} shows that lowering the speaking rate to 0.7 has the best performance for label 1 while the speaking rate of 0.8 works the best for label 2. Table \ref{tab:expB} shows one example which has high speaking rate (14 words in 3 seconds), it revels that baseline E2E model fail to recognize some words when the speaking rate is high. 


\subsection{Data augmentation: 3 way speed-perturbed (3W) and monolingual data (F)}

Table \ref{tab:B} shows that 3 way speed-perturbed (E) improves the performance significantly on both label 1 and label 2 data. Again, E2E+3W performs better with label 2 than label 1.
In order to observe the effect of adding different amount of monolingual data to the ASR performance, we create four different mixed datasets 
Table \ref{tab:F} shows the performance on SEAME test sets (without non language symbols) of systems trained with each of the four datasets, separately. It also presents the performance on ENG, MAN, and CS utterances. The system trained with F1 (adding 100 hours of Mandarin data) mostly improves the performance on the MAN speech. The one trained with F2 (adding 100 hours of English data) improves the performance on the ENG speech and interestingly also MAN and CS speech. When trained with F4 (adding 1000 hours of Mandarin and English data), they get worse MER than F3 (adding 200 hours of Mandarin and English data) because monolingual data becomes dominant in the train set. However, F4 has better performance on monolingual benchmarks (the system trained with F4 has 12.6\% WER on WSJ test and 10.1\% CER on AISHELL-1 test set, whereas the system trained with F3 has 26\% WER and 35\% CER on them). Note that the baseline system which trained only with SEAME data has over 100\% WER and CER on both monolingual benchmarks. The results indicate that optimizing the performance on both CS and monolingual test sets is an important trade-off which needs future investigations.\\ 

\tabcolsep=0.11cm
\begin{table}
\scriptsize
  \caption{\scriptsize The MER(\%) on SEAME test set of baseline and E2E+F approach with label 1}
  \label{tab:F}
  \centering
  \begin{tabular}{ |l |c c | ccc | ccc|}
    \hline
    &\multicolumn{2}{c|}{\textbf{No nlsyms}}&\multicolumn{3}{c|}{\textbf{dev}}&\multicolumn{3}{c|}{\textbf{eval}}\\
    \hline
    \textbf{} & \textbf{dev} & \textbf{eval} & \textbf{ENG} & \textbf{MAN}& \textbf{CS} & \textbf{ENG} & \textbf{MAN} & \textbf{CS}\\
    \hline
    E2E & 39.8 & 31.0 & 58.5 & 31.0 & 38.8 &45.9 & 28.6 & 30.2\\
    E2E+F1 & 39.5 & 30.1 & 61.1 & \textbf{29.5} & 38.4 & 46.0 & \textbf{27.3} & 29.2 \\
    E2E+F2 & 37.7  &  29.9 & \textbf{56.0} & \textbf{30.1} & \textbf{36.6} & \textbf{43.3} & \textbf{28.3} & \textbf{29.0}\\
    E2E+F3 & \textbf{37.5} & \textbf{28.8} & \textbf{56.5} & \textbf{28.6} & \textbf{36.5} & \textbf{43.2} & \textbf{26.5} & \textbf{28.0}\\
    E2E+F4 & 39.2  & 30.7 & 62.8 & 31.6 & 38.2 &46.9 & 29.4 & 29.5\\
    \hline
  \end{tabular}
\end{table}

\subsection{To combine all the approaches?}
\tabcolsep=0.11cm
\begin{table}[b]
\scriptsize
  \caption{\scriptsize The MER(\%) on SEAME test set of baseline and possible combinations with label 1.}
  \label{tab:combined}
  \centering
  \begin{tabular}{| l| c c |c c| }
    \hline
    &&&\multicolumn{2}{c|}{\textbf{No nlsyms}}\\
    \textbf{Systems} & \textbf{dev} & \textbf{eval} & \textbf{dev} & \textbf{eval} \\
    \hline
    E2ELD+SW     & 37.0  & 29.9 & 36.7  & 29.1 \\
    E2ELD+SL     & 36.2 & 29.7 & 35.8 & 28.9  \\
    E2ELD+3W     & 34.3 & 27.4 & 33.9 & 26.5\\
    E2ESW+SL     & 34.4 & 27.8 & 34.0 & 27.0\\
    E2ESW+3W     & 32.5 & \textbf{25.9}  & 32.0 & \textbf{25.1} \\
    E2E+SL+3W     & 32.8 & 26.1 & 32.3 & 25.2\\
    E2ESW+SL+3W   & \textbf{31.9} & 26.0 & \textbf{31.5} & \textbf{25.1}\\
    \hline
    E2ELD+3W+F3 & 34.0  & 26.8 & 33.5  & 25.9 \\
    E2ESW+SL+F3 & 32.8  & 26.4 & 32.3  & 25.5  \\
    E2ESW+3W+F3 & \textbf{31.4}  & \textbf{25.0} & \textbf{30.8}  & \textbf{24.2} \\
    \hline
  \end{tabular}
\end{table}
Note that all the combined systems are trained with label 1 data because we want to firstly find the best combined system, then apply it with label 2 data. Table \ref{tab:combined} shows that all the combinations except E2ELD+SW improve the MER and especially the combinations involving E2ESW, SL and 3W improve the most. The reasons why E2ELD+SW performs worst could lie in the fact that by introducing English subword containing 3 to 4 phones, it is much harder for the system to estimate the language identification.
Overall, adding monolingual data improves the performance of all the combinations. E2ESW+3W+F3 achieves the best performance with 25.0\% MER on the SEAME evaluation set. As mentioned before, we apply this best combination with label 2 data and achieve 23.7\% MER on the SEAME evaluation set.

\subsection{Language model fusion methods}
Results in Table \ref{tab:seametest} show the comparison between the baseline models and the best improved models without an external language model. For label 1 and label 2, the improved models achieve up to 35\% relative performance to the baseline models. Again, the model trained with label 2 data has the lowest MER. The second row in the Table shows the comparisons between the state-of-the-art TDNN-HMM \cite{b9} which applies i-vector and 3-way data perturbation techniques followed by a Kaldi chain recipe \cite{b40}. Kaldi exploits a bilingual pronunciation dictionary, which does not contain the pronunciation of discourse particles, hesitations' nor nonlinguistic signals, to train the TDNN-HMM chain model and integrates the language model using SF. The best improved model with SF and CF outperforms TDNN-HMM chain model with SF. 

Table \ref{tab:SFCF1} shows how our external language models can improve the best E2E model. As mentioned before, E2ESW+3W+F3 is not trained with the language identification loss function since subword modeling will harm the performance of the language identification. Therefore, it sometimes misrecognizes the English signal as a Mandarin signal. For example, it recognizes 'take' as \begin{CJK*}{UTF8}{gbsn}'带'\end{CJK*}. The external language model can help to increase the score of 'take initiative' in order to output the correct sentence. In this case, the models using shallow fusion or cold fusion to inject language model knowledge (E2ESW+3W+F3+SF and E2ESW+3W+F3+CF) do not make a mistake of recognizing languages.

Note that 'so' and \begin{CJK*}{UTF8}{gbsn}'所以'\end{CJK*} have similar pronunciation and meaning. The interesting example in Table \ref{tab:SFCF1} leads to the question of whether Mixed Error Rate is always reliable metrics in the context of CS speech recognition. From the perspective of automatic evaluation, E2ESW+3W+F3+CF performs worse than E2ESW+3W+F3+SF in this case since E2ESW+3W+F3+CF has longer Levenshtein distance (one substitution plus one insertion) to the ground truth sentence than E2ESW+3W+F3+SF's distance (one substitution). However from the perspective of human evaluation, E2ESW+3W+F3+CF performs better than E2ESW+3W+F3+SF with regard to the completion and meaning of entire sentence.

\begin{table}[t]
\scriptsize
  \caption{\scriptsize The MER(\%) on SEAME test set of baseline, improved combined model w/o or w/ LM and conventional TDNN-HMM hybrid system}
  \label{tab:seametest}
  \centering
   \begin{tabular}{|l| c| c| c c |c c|}
    \hline
    &\textbf{LM}&\textbf{label}&&&\multicolumn{2}{c|}{\textbf{No nlsyms}}\\
    \textbf{Systems}&\textbf{fusion} & \textbf{type}&\textbf{dev} & \textbf{eval} & \textbf{dev} & \textbf{eval} \\
    \hline
    E2E         &  & 1 &39.8 & 31.7 & 39.8 & 31.0 \\
    E2ESW+3W+F3 &  & 1 & 31.4  & 25.0 & 30.8 & 24.2\\
    E2E         &  & 2 &40.3 & 31.6 & 41.1 & 31.4 \\
    E2ESW+3W+F3 &  & 2 &\textbf{30.8}  & \textbf{23.7} & \textbf{30.7} & \textbf{23.0} \\
    \hline
    TDNN-HMM    &SF& 2 &35.9 &30.7& 35.3 & 29.7\\
    E2ESW+3W+F3 &SF &2& \textbf{29.8} & \textbf{22.8} & \textbf{29.7}  & \textbf{22.0} \\
    E2ESW+3W+F3 &CF &2& 29.9 & 23.0 & 29.7  & 22.2 \\
  \hline
  \end{tabular}
\end{table}

\begin{table}[!htbp]
\begin{CJK*}{UTF8}{gbsn}
\scriptsize
  \caption{\scriptsize The hypothesis of the best improved model w/ or w/o LM for utterances in eval set}
  \label{tab:SFCF1}
  \begin{center}
  \begin{tabular}{|l|l|}
  \hline
  {\bf Systems} &{\bf Hypothesis}\\
  \hline
    Ground-truth & then \hspace{1mm}你 \hspace{1mm}不 \hspace{1mm}可 \hspace{1mm}以 \hspace{1mm}take initiative\hspace{1mm} 去\hspace{1mm}讲 \hspace{1mm}么\\
    E2ESW+3W+F3  & then \hspace{1mm}你 \hspace{1mm}不 \hspace{1mm}可 \hspace{1mm}以 \hspace{1mm}\textcolor{red}{带} \hspace{1mm}initiative\hspace{2mm}\textcolor{red}{就} \hspace{1mm}讲\\
    E2ESW+3W+F3+SF & then \hspace{1mm}你 \hspace{1mm}不 \hspace{1mm}可 \hspace{1mm}以 \hspace{1mm}take initiative\hspace{1mm} 去\hspace{1mm}讲 \hspace{1mm}么 \\
    E2ESW+3W+F3+CF & then \hspace{1mm}你 \hspace{1mm}不 \hspace{1mm}可 \hspace{1mm}以 \hspace{1mm}take initiative\hspace{1mm} 去\hspace{1mm}讲\\
     \hline
    Ground-truth & 所\hspace{1mm}以 \hspace{1mm}我 \hspace{1mm}就 \hspace{1mm}去 \hspace{1mm}apply job\\
    E2ESW+3W+F3 & 所\hspace{1mm}以 \hspace{1mm}我 \hspace{1mm}就 \hspace{1mm}去 \hspace{1mm}apply job\\
    E2ESW+3W+F3+SF & 所\hspace{1mm}以 \hspace{1mm}我 \hspace{1mm}就 \hspace{1mm}去 \hspace{1mm}\textcolor{red}{ply} job\\
    E2ESW+3W+F3+CF & \textcolor{red}{so}\hspace{5mm}我 \hspace{1mm}就 \hspace{1mm}去 \hspace{1mm}apply job\\
  \hline
  \end{tabular}
  \end{center}
\end{CJK*}
\end{table}


\section{Conclusions}
We analyze several subproblems of Mandarin-English CS speech based on SEAME dataset, and provide solutions to each subproblem within the E2E ASR framework. 
We explore different combinations of the proposed solutions in order to reach the optimal ASR performance. The experimental results reveal that each solution improves the MER little by little, and the appropriate combination achieves great improvement (up to 35\% relatively). Our best combined system with an external language outperforms the baseline and the state-of-the-art hybrid system (TDNN-HMM). 




\begin{thebibliography}{1}

\bibitem{b1} P. Auer, ``Code-switching in conversation: language, interaction and identity,'' The Modern Language Review, vol. 95, 2000.
\bibitem{b2} S. Poplack, ``Sometimes I’ll start a sentence in Spanish y termino en espanol: toward a typology of code-switching 1,'' Linguistics, vol. 18, no. 7–8, 1980, pp. 581–618.
\bibitem{b3} D.-C.~Lyu, T.~P. Tan, C.~E. Siong, and H.~Li, ``An analysis of a Mandarin-English code-switching speech corpus: SEAME,'' in Proc. of INTERSPEECH, 2010.
\bibitem{b4} D.-C. Lyu, T. P. Tan, C. E. Siong, and H. Li, ``SEAME: a Mandarin-English code-switching speech corpus in south-east Asia,'' in Proc. of INTERSPEECH, 2010.
\bibitem{b5} N. T. Vu et al., ``A first speech recognition system for Mandarin-English code-switch conversational speech,'' in Proc. of ICASSP, 2012.
\bibitem{b6} G. E. Dahl, D. Yu, L. Deng, and A. Acero, ``Context-dependent pre-trained deep neural networks for large-vocabulary speech recognition,'' IEEE/ACMTrans.Audio, Speech, Language Process, vol. 20, 2012, pp. 30–42.
\bibitem{b7} G. Hinton et al., ``Deep neural networks for acoustic modeling in speech recognition: the shared views of four research groups,'' IEEE Signal Processing Magazine, vol. 29, 2012, pp. 82–97.
\bibitem{b9} V. Peddinti, D. Povey, and S. Khudanpur, ``A time delay neural network architecture for efficient modeling of long temporal contexts,'' in Proc. of INTERSPEECH, 2015.
\bibitem{b10} A. Graves and N. Jaitly, ``Towards end-to-end speech recognition with recurrent neural networks,'' in Proc. of ICML, 2014.
\bibitem{b14} A. L. Maas, Z. Xie, D. Jurafsky, and A. Y. Ng, ``Lexicon-free conversational speech recognition with neural networks,'' in Proc. of NAACL, 2015.
\bibitem{b15} D. Bahdanau, J. Chorowski, D. Serdyuk, P. Brakel, and Y. Bengio, ``End-to-end attention-based large vocabulary speech recognition,'' in Proc. of ICASSP, 2016.
\bibitem{b18} G. Klein, Y. Kim, Y. Deng, J. Senellart, and A. M. Rush, ``Opennmt: Open-source toolkit for neural machine translation,'' in EAMT User Studies and Project/Product Descriptions, 2017.
\bibitem{b20} T. Hori, S. Watanabe, and J. Hershey, ``Joint CTC/attention decoding for end-to-end speech recognition,'' in Proc. of ACL, 2017.
\bibitem{b21} F. Zhang and P. Yin, ``A study of pronunciation problems of English learners in China,'' in Asian Social Science, vol. 5(6), 2009.
\bibitem{b22} P. Roach, ``English phonetics and phonology: a practical course,'' in Cambridge University Press, 2000.
\bibitem{b24} Mandarin-English code-switching in south-east Asia. [Online]. Available: \url{https://catalog.ldc.upenn.edu/LDC2015S04}. [Accessed: May- 2019].
\bibitem{b25} P. Auer, ``From code switching via language mixing to fused lects toward a dynamic typology of bilingual speech,'' in International Journal of Bilingualism, vol. 3, 1999, pp. 309–332.
\bibitem{b28} S. Watanabe et al., ``ESPnet: end-to-end speech processing toolkit,'' in Proc. of INTERSPEECH, 2018.
\bibitem{b29} T. Mikolov, S. Kombrink, L. Burget, J. H. \v Cernock\'y, and S. Khudanpur, ``Extensions of recurrent neural network language model,'' in Proc. of ICASSP, 2011.
\bibitem{b30} Z. Xiao, Z. Ou, W. Chu, and H. Lin, ``Hybrid CTC-attention based end-to-end speech recognition using subword units,'' in Proc. of ISCSL, 2018.
\bibitem{b42} Z. Zeng et al., ``On the End-to-End Solution to Mandarin-English Code-switching Speech Recognition.'' in Proc. of INTERSPEECH, 2019.
\bibitem{b31} T. Ko1, V. Peddinti, D. Povey, and S. Khudanpur, ``Audio augmentation for speech recognition,'' in Proc. of INTERSPEECH, 2015.
\bibitem{b41} S. Toshniwal, A. Kannan, C.-C. Chiu, Y. Wu, T. N Sainath, K. Livescu, ``A comparison of techniques for language model integration in encoder-decoder speech recognition,'' in Proc. of SLT, 2018.
\bibitem{b34} H. Bu, J. Du, X. Na, B. Wu, and H. Zheng, ``AIShell-1: an open-source Mandarin speech corpus and a speech recognition baseline,'' in Proc. of Oriental COCOSDA, 2017.
\bibitem{b35} THCHS-30. [Online]. Available: \url{http://arxiv.org/abs/1512.01882}. [Accessed: May- 2019].
\bibitem{b36} Free ST Chinese Mandarin corpus. [Online]. Available: \url{http://www.openslr.org/38/}  [Accessed: May- 2019].
\bibitem{b37} V. Panayotov, G. Chen, D. Povey, and S. Khudanpur, “Librispeech: an ASR corpus based on public domain audio books,” in Proc. of ICASSP, 2015.
\bibitem{b38} Mozilla common voice. [Online]. Available: \url{https://voice.mozilla.org/en}. [Accessed: May- 2019].
\bibitem{b39} TedXSingapore. [Online]. Available: \url{ https://www.ted.com/tedx/events/25530}. [Accessed: May- 2019].
\bibitem{b40} D. Povey et al., ``The Kaldi speech recognition toolkit,'' in Proc. of ASRU, 2011.
\end{thebibliography}
%

\end{document}